\pdfoutput=1
\documentclass[final]{cvpr}

\usepackage{times}
\usepackage{epsfig}
\usepackage{graphicx}
\usepackage{amsmath}
\usepackage{amssymb}
\usepackage{multirow}
\usepackage{xcolor}
\usepackage{hyphenat}
\usepackage{enumitem,kantlipsum}
\usepackage{url}            %
\usepackage{booktabs}       %
\usepackage{amsfonts}       %
\usepackage{nicefrac}       %
\usepackage{microtype}      %
\usepackage{amssymb}
\usepackage{amsmath}
\usepackage{subfigure}
\usepackage{lipsum,multicol}
\usepackage{multirow}
\usepackage{array}
\usepackage{amsthm}

\DeclareMathOperator*{\argmin}{\arg\!\min}
\newcommand{\bx}{\mathbf{x}}

\newcommand{\by}{\mathbf{y}}

\newcommand{\bh}{\mathbf{h}}
\newcommand{\bz}{\mathbf{z}}

\newcommand{\CD}{\mathcal{D}}
\newcommand{\CW}{\mathcal{W}}

\newcommand{\bw}{\mathbf{w}}

\newcommand{\train}{\text{train}}

\newcommand{\ce}{\text{CCE}}

\usepackage[pagebackref=true,breaklinks=true,colorlinks,bookmarks=false]{hyperref}

\def\cvprPaperID{52} %
\def\confYear{CVPR 2021}
\begin{document}

\title{Contrastive Learning Improves Model Robustness Under Label Noise}

\author{Aritra Ghosh \qquad \qquad Andrew Lan\\
University of Massachusetts Amherst\\
{\tt\small \{arighosh,andrewlan\}@cs.umass.edu}
}

\maketitle

\begin{abstract}
Deep neural network-based classifiers trained with the categorical cross-entropy (CCE) loss are sensitive to label noise in the training data. One common type of method that can mitigate the impact of label noise can be viewed as supervised robust methods; one can simply replace the CCE loss with a loss that is robust to label noise, or re-weight training samples and down-weight those with higher loss values. Recently, another type of method using semi-supervised learning (SSL) has been proposed, which augments these supervised robust methods to exploit (possibly) noisy samples more effectively. Although supervised robust methods perform well across different data types, they have been shown to be inferior to the SSL methods on image classification tasks under label noise. Therefore, it remains to be seen that whether these supervised robust methods can also perform well if they can utilize the unlabeled samples more effectively. In this paper, we show that by initializing supervised robust methods using representations learned through contrastive learning leads to significantly improved performance under label noise. Surprisingly, even the simplest method (training a classifier with the CCE loss) can outperform the state-of-the-art SSL method by more than 50\% under high label noise when initialized with contrastive learning. Our implementation will be publicly available at {\url{https://github.com/arghosh/noisy_label_pretrain}}.
\end{abstract}

\vspace{-0.5cm}
\section{Learning under Label Noise}
In standard classification tasks, we are given a dataset $\CD=\{(\bx_i,\by_i)\}_{i=1}^N$ where  $\bx_i$ is the feature vector of the  $i^{\text{th}}$ sample (image) and $\by_i\in \{0,1\}^K$ is the class label vector with $K$ total classes. We minimize the following empirical risk minimization (ERM) objective, 
\vspace{-0.15in}
\begin{equation}
\min_{\bw}\frac{1}{N}  \sum_{i=1}^N \ell_{\ce}(\by_i, f(\bx_i;\bw)),
\label{eq:erm}
\vspace{-0.15in}
\end{equation}
where $f(\cdot; \bw)$ is a deep neural network (DNN)-based classifier 
with parameters $\bw$ and $\ell_{\ce}$ is the categorical cross-entropy (CCE) loss. However, real-world datasets often contain noisy labels; i.e., $\by_i$ can be corrupted. DNNs are sensitive to label noise when they are trained with the CCE loss, which reduces %
their ability to generalize to clean dataset.

Most of the early works on learning under label noise can be called %
as \emph{supervised robust methods} %
and they are equally applicable to image, text, or any other data types. A general trick
 to mitigate the impact of 
label noise is to replace the CCE loss function $\ell_{\ce}$ with a loss that is more robust to label noise 
in Eq.~\ref{eq:erm} \cite{generalized-ce,peer,normalized-loss,ghosh2015,ghosh2017-dt,ghosh2017,unhinged,peer,ldmi}. In \cite{ghosh2017},  the authors show that a loss function $\ell$ is robust to uniform label noise if it
satisfies the condition $\sum_{k=1}^K\ell(k,f(\bx_i;\bw))=C$ for some constant $C$. The mean absolute error (MAE) loss satisfies this symmetric condition; however, the MAE loss is difficult to optimize under the ERM objective with DNNs. Several loss functions have been proposed that offer model robustness under label noise and  they are easier to optimize compared to the MAE loss \cite{peer,ldmi,generalized-ce,normalized-loss}. For example, the generalized cross-entropy loss $L_q$ ($q\in (0,1]$ is a hyper-parameter) is defined as \cite{generalized-ce}
    $L_q(\by,f(\bx;\bw)) = \frac{1-\by^{\intercal}f(\bx;\bw)^q }{q}$.
The $L_q$ loss is equivalent to the CCE loss when $q\!\rightarrow\! 0$ and is equivalent to the MAE loss when $q\!=\!1$. However, these robust loss functions do not perform well on large image datasets. 

Another common strategy for learning under label noise is to separate out the noisy samples from the clean samples or re-weight the training samples and stick with the CCE loss; we can simply change the objective as%
\vspace{-0.2cm}
\begin{align}
\min_{\bw}\frac{1}{N} \sum_{i=1}^N \CW(\bx_i,\by_i)  \ell_{\ce}(\by_i, f(\bx_i;\bw)),\label{eq:weight-erm}
\vspace{-0.2cm}
\end{align}
where $\CW(\bx_i,\by_i)\in [0,1]$ is the assigned weight for the training sample $(\bx_i,\by_i)$. 
A common heuristic, studied in earlier research, is that %
noisy samples have higher loss values compared to the clean samples \cite{identifying-mislabeled,identifying}. 
Many recent methods apply this idea to filter out or lower %
weights to possibly noisy samples \cite{s-model,co-teaching,joint,webly,co-teaching,mentornet,joint,veit2017learning,yuan2018iterative,focal,reed,masking}. Instead of filtering samples based on the loss value, a more principled way is to \emph{learn} a weighting \emph{function} $\CW(\ell(\by_i, f(\bx_i;\bw));\theta)$ in a data-driven manner where the function takes the loss value as the input. 
Meta-learning-based methods have been particularly useful to learn a weighting function \cite{mwnet,robust-mwnet,l2rw,l2l-noise}. As an example, 
Meta-Weight Network (MWNet) \cite{mwnet} learns a weighting function $\CW$ with parameter $\theta$ using a small number of clean  validation samples in a bilevel setup \cite{bilevel}. The objective is to learn the optimal weighting function $\CW(\ell(\cdot,\cdot); \theta^{\ast})$ such that the optimal classifier parameters $\bw^{\ast}(\theta^{\ast})$ on the training samples (train), obtained from Eq.~\ref{eq:weight-erm}, optimizes the ERM objective on the clean validation samples (val). The bilevel optimization problem can be written as %
\vspace{-0.2cm}
\begin{align*}
&\quad\quad\quad\quad\quad\quad\min_{\theta} \sum_{j\in \text{val}} \ell_{}\Big(\by_j,f(\bx_j;\bw^{\ast}(\theta))\Big)\\
&\mbox{ s.t. } \bw^{\ast}\!(\theta)\!=\!\!\argmin_{\bw}\! \!\!\sum_{i\in \train}\!\!\! \CW\!\Big(\!\ell_{}\!(\by_i,\!f(\bx_i;\bw));\theta\!\Big)
\!\ell_{}\!(\by_i,\!f(\bx_i;\bw)\!). 
\vspace{-0.2cm}
\end{align*}
These \emph{supervised robust methods} perform well across many data types. %

Recently, many semi-supervised learning (SSL) methods have been proposed for image datasets to mitigate the impact of label noise. %
SSL methods aim to improve the performance of a DNN classifier by exploiting unlabeled data \cite{mixmatch}. Common tricks in SSL methods include %
using a consistency regularization loss to encourage %
the classifier to have similar predictions for %
an image $\bx_i$ and the augmented view of the image $\text{Aug}(\bx_i)$, an entropy minimization objective to promote high confidence predictions, and a label guessing method to produce a good guess from many augmentations of the same image \cite{mixmatch,mixup}. 
DivideMix, an SSL method, divides the training dataset into the clean (labeled) and noisy (unlabeled) parts using the observation that noisy samples tend to have a higher loss value \cite{dividemix}. These SSL methods have been shown to be superior to the supervised robust methods on image datasets \cite{elr,dividemix}.

\vspace{-0.1cm}
\subsection{Contributions}
\vspace{-0.1cm}
We observe that SSL methods for label noise can use unlabeled (noisy) samples effectively to improve 
their representation learning capability. 
Consequently, prior supervised robust learning methods suffer a significant drop in performance compared to the SSL methods on image datasets.
Hence, we ask the following question:
\vspace{-0.2cm}
\begin{itemize}
[leftmargin=*]
\item Is the performance drop of supervised robust methods caused by label noise or the impaired representation %
learned using fewer clean samples?
\vspace{-0.2cm}
\end{itemize}
Thus, we study the effect of fine-tuning these supervised robust methods after initializing them with \emph{good} representations learned by a self-supervised method. Contrastive learning has emerged as a key method for self-supervised learning 
from visual data; the general idea is to learn good visual representations of images through comparing and contrasting different views of an original image under various data augmentation operations \cite{simclr,simclrv2}. We find that the supervised robust methods work remarkably well when they are initialized with the contrastive representation learning model. Surprisingly, we notice that even using the (most sensitive) CCE loss can outperform state-of-the-art SSL methods under high %
label noise. Moreover, we observe that the generalized cross-entropy loss \cite{generalized-ce} can retain good performance even under 95\% uniform label noise on the CIFAR-100 dataset whereas training with a random initializer does not outperform a random model. These observations suggest that the drop in performance for the supervised robust methods is due to the lack of good visual representations. We use one representative method from each of the two major paradigms we described for the supervised robust methods (the $L_q$ loss for the loss correction approach and the MWNet method for the sample re-weighting strategy) 
to illustrate the benefits of fine-tuning representations learned through contrastive learning with a classification task under label noise.
\vspace{-0.1cm}
\subsection{Related Works}
\vspace{-0.1cm}
The idea of using a pre-trained model initializer or self-supervised learning is not new in label noise research. 
In \cite{self-supervision}, the authors use auxiliary tasks, such as rotation prediction, to improve model robustness under label noise.
In \cite{pretraining}, the authors propose to use a pre-trained Imagenet classifier to improve model robustness. These methods lead to improved performance under high label noise, adversarial perturbation, class imbalance conditions, and on out-of-distribution detection tasks. However, they require a larger similar dataset where label noise is not present %
or need to use an auxiliary loss from the self-supervised tasks in addition to the classification task. In contrast, our work does not propose any additional auxiliary tasks or require any larger datasets. We learn the contrastive model for visual representations from the same dataset as the classification task. This is helpful when the classification task uses datasets (e.g., in medical imaging datasets) that are very different than the commonly used large-scale image datasets (e.g., the ImageNet dataset). 
The most related work is \cite{c2d}, which uses a contrastive model to improve the DivideMix algorithm. However, we show that a self-supervised contrastive learning model 
initializer can improve model robustness under label noise for many supervised robust methods.

\vspace{-0.1cm}
\subsection{Methodology}
\vspace{-0.1cm}
We will use the SimCLR framework for contrastive learning \cite{simclr,simclrv2}; however, other visual representation learning methods (including other contrastive learning methods) can also potentially improve model robustness under label noise.
We use a base encoder $\hat{f}(\cdot)$ (ResNet-50 in this paper) to encode each image ${\bx}_i$ to $\bh_i=\hat{f}({\bx}_i)$, and a two-layer multi-layer perceptron  $g(\cdot)$ as the projection head to project into a fixed dimension embedding $\bz_i=g(\bh_i)$. %
Using $M$ images and two augmentations for each image, we construct a dataset of $2M$ images $\{\bx_{i,0}, \bx_{i,1}\}_{i=1}^M$ and project them into $\{\bz_{i,0}, \bz_{i,1}\}_{i=1}^M$ using the base encoder and the projection head. The final objective in the SimCLR framework is defined as%
\vspace{-0.1cm}
\begin{equation*}
    \sum_{i=1}^M\!\sum_{j=0}^1 \!-\!\log\!\frac{\exp{(\text{sim}(\bz_{i,j},\bz_{i,j+1\%2})/\tau)}}{\!-\!\exp{(1/\tau)}\!+\!\sum_{k=1,l=0}^{k=M,l=1}\!\exp\!{(\text{sim}(\bz_{i,j}\!,\!\bz_{k,l})\!/\!\tau\!)}}\!,
    \vspace{-0.1cm}
\end{equation*}
where $\tau$ is the temperature parameter, and $\text{sim}(\bz_i,\bz_j)$ is the normalized cosine similarity $\frac{\bz_i^{\intercal}\bz_j}{||\bz_i||||\bz_j||}$. 
We use the same dataset $\CD=\{\bx_i,\by_i\}_{i=1}^N$ to learn the SimCLR encoder $\hat{f}(\cdot)$. 
The base encoder $\hat{f}$ does not contain the classification head (last output layer). For supervised robust methods, we use this encoder $\hat{f}$ to initialize the DNN classifier $f(\cdot;\bw)$ and we set the weights and biases of the classification head of $f(\cdot;\bw)$ to zero at initialization. 
Note that %
we fine-tune the final classifier $f(\cdot)$ for each method and do not keep the base encoder $\hat{f}(\cdot)$ fixed.

\begin{table*}[t]\centering
    \scalebox{0.64}{
    \begin{tabular}{ccc cccccccc| cccccccc}\toprule
    \multicolumn{3}{c}{Noise Rate (\%)}& $0$ & $20$ &$40$ &$50$ &$60$& $80$ & $90$ & $95$& $0$   &  $20$ &  $40$ &$50$ &$60$ & $80$ & $90$ & $95$\\
    \cline{1-19}        
 \multirow{ 1}{*}{Method} & \multirow{ 1}{*}{Arch} & \multirow{ 1}{*}{Initializer}    &  \multicolumn{8}{c|}{CIFAR-10} &  \multicolumn{8}{c}{CIFAR-100} \\
    \cline{1-19}
            
    CCE  \cite{dividemix}& \multirow{4}{*}{PRN-18} & \multirow{4}{*}{No} &  \multirow{4}{*}{95.11} &86.8&- & 79.4&- & 62.9 & 42.7& -&  \multirow{4}{*}{74.4}  &62.0&-& 46.7&- & 19.9 & 10.1  &-\\
    MLNT \cite{l2l-noise,dividemix} &  &  &  &92.9 &-& 89.3 &-& 77.4 & 58.7 &-  && 68.5&- &59.2&- & 42.4 & 19.5 & -\\
    F-Correction \cite{forward,dividemix} &  && & 86.8 &-& 79.8 &-& 63.3 &42.9 & - && 61.5&- &46.6&- & 19.9 & 10.2\\
    M-Correction \cite{m-correction,dividemix}&  & & &94.0&- & 92.0 &-& 86.8 & 69.1 & - &&  73.9&- & 66.1&- & 48.2 & 24.3 &-\\
    \cline{2-19}
    Divide-Mix \cite{dividemix} & PRN-18 &No & {95.8}$^{\ast}$&  96.1 &-& 94.6 &-& {\bf 93.2} & 76 & -& {78.9}$^{\ast}$  &77.3 &-& 74.6 &-& 60.2 & 31.5 & -\\
    \cline{2-19}
    MWNet \cite{mwnet} &WRN-28-10 & No &95.60&92.45 & 89.27&87.49& 84.07 & 69.65 &25.8 & 18.49& 79.95&73.99& 67.73 &66.88&58.75&30.55 & 5.25&3.05 \\
    \cline{2-19}
    $L_{q}$ \cite{generalized-ce} & \multirow{2}{*}{RN-34} & \multirow{2}{*}{No} &\multirow{2}{*}{93.34} &89.83 & 87.13 &-& 82.54 & 64.07& - & -&\multirow{2}{*}{76.76} &66.81 &61.77 &-&53.16& 29.16 &-&-\\
     ELR \cite{elr} &  &  &&92.12& 91.43 & &88.87 & 80.69& -& -&&74.68  &68.43 & -& 60.05 & 30.27 & -&-\\
    \midrule
      CCE & \multirow{3}{*}{RN-50} & \multirow{3}{*}{INet32} &  \multirow{3}{*}{\bf 96.52} & 92.37 & -& 91.56& - & 83.34 & 62.66 & 39.09 &\multirow{3}{*}{\bf 81.51} & 70.26 & -& 65.76& -& 54.33 & 38.9 & 20.59 \\
    $L_q$ &  &  & & 94.07&- & 93.75&- & 90.56& 85.89 & 73.14 & & 77.22 &-& 69.87&- & 60.5 & 54.83 & 44.3\\
    MWNet &  & & & {\bf 97.33} & -& {\bf 96.17} &-& { 93.12} & {\bf 90.88} & 85.27 & & {\bf 82.85} &-& {\bf 80.28} &-&{\bf 71.29} & {\bf 58.21} & {44.62}\\
    \midrule
    CCE & \multirow{3}{*}{RN-50} & \multirow{3}{*}{SimCLR} & \multirow{3}{*}{94.59} & 93.29 & -& 91.96 & -& 88.75 & 82.9 & 66.07 & \multirow{3}{*}{75.36} & 71.98 & -& 67.89 &- &59.84& 52.11 & 39.57\\
    $L_q$ &  &  & & 94.02 &-& 92.94 &-&90.85 & 88.45 & 83.76 & & 73.33 & -&70.14 &-&63.26 &55.93 & {\bf 45.7} \\
    MWNet &  &  & & 93.88 &-& 92.92 &-&91.51 &90.19 & {\bf 87.23} & & 73.2 & -& 69.88 &-&64.05 & 57.6&{ 44.91}\\
      \bottomrule
        \end{tabular}
    }%
    \caption{
    Test accuracy (\%) for various methods on the CIFAR datasets under symmetric label noise. We re-use results for the `No' initializer cases from their respective papers. We re-run public implementation of MWNet \cite{mwnet} for some (missing) noise levels.  The test accuracy under 0\% label noise refers to the accuracy obtained from minimizing the ERM objective with the CCE loss except for DivideMix ($^{\ast}$) for which test accuracy is obtained from training the MixUp objective with Preactivated ResNet-18 (PRN-18) \cite{mixup}. We bold performance for the best method under each noise settings.
    }
    \label{tab:sym}
    \vspace{-0.4cm}
\end{table*}

\begin{table*}[t]\centering
    \scalebox{0.67}{
    \begin{tabular}{ccc cccc| cccc}\toprule
    \multicolumn{3}{c}{Noise Rate (\%)}& $0$ &  $20$ &$30$ &$40$ &  $0$ &$20$ &  $30$ &$40$ \\
    \cline{1-11}
 \multirow{ 1}{*}{Method} & \multirow{ 1}{*}{Arch} & \multirow{ 1}{*}{Initializer}    &  \multicolumn{4}{c|}{CIFAR-10} &  \multicolumn{4}{c}{CIFAR-100} \\
    \cline{1-11}      
    CCE \cite{generalized-ce} & \multirow{4}{*}{RN-34} & \multirow{4}{*}{No}   &\multirow{4}{*}{93.34} &88.59 & 86.14&  80.11&  \multirow{4}{*}{76.76}&  59.20& 51.40 & 42.74\\
    { F-correction\cite{forward,generalized-ce}} & &  & &90.35& 89.25 & 88.12 && 71.08& 70.76& 70.82\\
    $L_q$\cite{generalized-ce} &  && &89.33& 85.45 & 76.74& &66.59& 61.45& 47.22\\
    ELR \cite{elr} &  &   & &93.28 & 92.70 & 90.35&  &74.20 & 74.02 & {\bf 73.73}\\
    \cline{2-11}
    MWNet \cite{mwnet} & WRN-28-10 & No  & 95.60&93.14 & 91.45 &89.71 & 79.95 &71.55 &66.07  &56.05 \\
    \midrule
      CCE & \multirow{3}{*}{RN-50} & \multirow{3}{*}{INet32} &\multirow{3}{*}{\bf 96.52} &92.93 & 91.78 & 90.22 & \multirow{3}{*}{\bf 81.51}& 69.76 & 62.41 & 52.4\\
    $L_q$ &  &  & & 93.77 & 93.23 & 90.24 & & 71.63 & 67.29 & 59.29\\
    MWNet &  &  & & {\bf 96.85} & {\bf 95.9} & {\bf 94.99} & & {\bf 80.74} & {\bf 77.36} & 72.93\\
    \midrule
    CCE & \multirow{3}{*}{RN-50} & \multirow{3}{*}{SimCLR} & \multirow{3}{*}{94.59}& 93.3 & 92.13 & 88.38 & \multirow{3}{*}{75.36}& 69.63 & 63.91 & 54.28\\
    $L_q$ &  &  & & 93.54 & 92.69 & 90.27 & & 71.26 & 68.04 & 59.26\\
    MWNet &  &  & & 93.67 & 93.18 & 92.59 & & 72.17 & 69.86 & 64.92\\
      \bottomrule
        \end{tabular}
    }%
    \vspace{0.03cm}
    \caption{
    Test accuracy (\%) for various methods on the CIFAR datasets under asymmetric label noise. We re-use results for the `No' initializer cases from their respective papers except for MWNet for which we run their public implementation with asymmetric noise.  %
    }
    \label{tab:asym}
   \vspace{-0.4cm}
\end{table*}
\begin{table}[t]\centering
    \scalebox{0.65}{
    \begin{tabular}{ccc}\toprule
Method & Initializer & Accuracy\\
\midrule
CCE \cite{dividemix} & \multirow{6}{*}{ImageNet} & 68.94 \\
F-Correction \cite{forward} &  & 69.84\\
MLNT \cite{l2l-noise} &  & 73.47\\
DivideMix \cite{dividemix} & & 74.76\\
ELR \cite{elr}&  & 72.87\\
ELR+ \cite{elr}&  & 74.81\\
\midrule
CCE &\multirow{3}{*}{ SimCLR} & 73.27\\
$L_q$ &  & 73.35\\
MAE &  & 73.36\\
      \bottomrule
        \end{tabular}
    }%
    \caption{Test accuracy (\%) for various methods on  Clothing1M. We use results for the ImageNet initializer from their respective papers. 
    }
    \label{tab:clothing}
    \vspace{-0.4cm}
\end{table}

\vspace{-0.1cm}
\section{Experimental Results}
\vspace{-0.1cm}

{\bf Datasets and Experimental Setup:} We demonstrate the efficacy of our proposed approach on CIFAR-10, CIFAR-100, and Clothing1M datasets. Unless otherwise specified, we use ResNet-50 (RN-50) as the classifier; for CIFAR datasets, we adopt the common practice of replacing the first convolutional layer of kernel size 7, stride 2 with a convolutional layer of kernel size 3 and stride 1 and removing the first max-pool operation in RN-50 \cite{simclr}. 

CIFAR-10 and CIFAR-100 datasets contain 50k training samples and 10k test samples; label noise is introduced synthetically on the training samples. We keep 1000 clean training samples for validation purposes. We experiment with symmetric noise and asymmetric noise. Under symmetric noise, the true class label is changed to any of the class labels (including the true label) whereas, under asymmetric noise, the true class label is changed to a similar class label. We use the exact same setup of \cite{generalized-ce,forward} for introducing asymmetric noise. For CIFAR-10, the class mappings are TRUCK $\rightarrow$ AUTOMOBILE, BIRD $\rightarrow$ AIRPLANE, DEER $\rightarrow$ HORSE, CAT $\leftrightarrow$ DOG. For CIFAR-100, the class mappings are generated from the next class in that group (where 100 classes are categorized into 20 groups of 5 classes). 

Clothing1M dataset is real-world datasets consisting of 1M training samples; labels are generated from surrounding text in an online shopping website \cite{xiao2015learning}. Clothing1M dataset contains around 38\% noisy samples \cite{c2d} and we do not introduce any additional noise on this dataset.

{\bf Pre-Training:} We compare the supervised robust methods using two initialization, namely the SimCLR initializer and the ImageNet pre-trained initializer \cite{pretraining}. 
To train the SimCLR encoder $\hat{f}(\cdot)$ and the projection head $g(\cdot)$, we use a batch size of 1024 (1200) and run for 1000 (300) epochs  with the LARS optimizer \cite{lars} on a single NVIDIA RTX8000 (12 NVIDIA M40) GPU(s) on the CIFAR-10/100 (Clothing1M) datasets. 
For the Clothing1M dataset, we use the standard pre-trained ImageNet RN-50 initialization. For the CIFAR datasets, we train a RN-50 classifier from scratch (with CIFAR changes in the first convolutional layer) on the ImageNet-$32\times32$ (INet32) dataset \cite{imagenet32} that achieves 43.67\% Top-1 validation accuracy. %

{\bf Methods:}
 We use the SimCLR RN-50 initializer for three methods: standard ERM training with the CCE loss, ERM training with the generalized cross-entropy loss $L_q$  (q=0.5 or 0.66), and MWNet. The value of q in $L_q$ loss is important only for a high rate of noise ($\geq 0.8$), where the $L_q$ loss with a large q (0.66) is difficult to optimize; however, for all other noise rates, a higher value of q leads to better performance. We fine-tune for 120 epochs on the CIFAR datasets with the SGD optimizer, learning rate of 0.01, momentum of 0.9, weight-decay of 0.0001, and a batch size of 100.
For other baseline methods, we use the results listed in their respective paper (or their public implementation). 
Note that 
different prior works use different architectures; thus, we also list the test accuracy from training on clean samples with that architecture (and initialization).  
For CIFAR datasets, we list the average accuracy from five runs of noisy label generation.  

We use the CCE loss, the $L_q$ loss (q=0.66), and the MAE loss with the SimCLR initializer on the Clothing1M dataset. We use the SGD optimizer, a batch size of 32, momentum of 0.9, and an initial learning rate of 0.001. Following \cite{elr,dividemix}, we randomly sample 4000x32 training samples in each epoch such that the total number of samples from each of the classes are equal. We fine-tune for 60 epochs and reduce the learning rate by a factor of 2 after every 10 epochs.

\vspace{-0.1cm}
\subsection{Results and Discussion}
\vspace{-0.1cm}
Table~\ref{tab:sym} lists classification performance on the test set under symmetric noise on the CIFAR datasets. The SimCLR initializer significantly improves performance for the CCE loss, the $L_q$ loss, and the MWNet method. 
Under 90\% label noise, the CCE loss has an accuracy of 42.7\% (10.1\%) with a random initializer and DivideMix has an accuracy of 93.2\% (31.5\%) on the CIFAR-10 (CIFAR-100) dataset. Under the same noise rate, the CCE loss with the SimCLR initializer has an accuracy of 82.9\% (52.11\%) on the CIFAR-10 (100) dataset which translates to a 9\% (65\%) gain compared to the state-of-the-art method DivideMix. 
Moreover, the SimCLR initializer beats these performances even further with the MWNet method and the $L_q$ loss. Under very high levels of label noise, MWNet and the $L_q$ loss are not able to learn anything useful with the standard random initializer. However, with the SimCLR initializer, these methods perform significantly better than the state-of-the-art method.

Table~\ref{tab:asym} lists the classification performance on the test set under asymmetric label noise on the CIFAR datasets. Similarly, we observe that the SimCLR initializer improves  %
model robustness under asymmetric label noise. However, supervised robust methods do not beat the prior-state-of-the-art method for the asymmetric noise case.%

Table~\ref{tab:clothing} lists the test performance on the Clothing1M dataset. Although the CCE loss, the $L_q$ loss, and the MAE loss do not outperform state-of-the-art methods with the SimCLR initializer, they perform remarkably well. 

We also observe that RN-50 pre-trained on the INet32 dataset also improves model robustness under label noise on the CIFAR datasets. Note that there is significant overlap between the classes of the INet32 dataset and the classes of the CIFAR datasets. On the CIFAR-100 dataset, the RN-50 pre-trained model (on INet32) significantly improves test accuracy to 81.51\% from fine-tuning compared to $\sim75\%$ with a random or the SimCLR initializer. 
However, the SimCLR initializer does not require such a considerable knowledge transfer from another larger dataset. Moreover, we observe that the drop in performance (w.r.t.\ the accuracy from the clean training samples) is significantly lower for the SimCLR initializer compared to the pre-trained ImageNet initializer. The pre-trained INet32 initializer seems to help more in the case of the asymmetric noise case; class overlap and label corruptions to a similar class might be a reason behind the improvement. In contrast, in the Clothing1M dataset, only two of the classes (out of 14) are present in the ImageNet dataset.
Consequently, the CCE loss with the SimCLR initializer improves the test performance by 6\% 
compared to the ImageNet pre-trained RN-50 initializer.

\vspace{-0.1cm}
\section{Conclusion}
\vspace{-0.1cm}
In this paper, we have shown that many of the supervised robust methods do not learn anything useful under high label noise rates. 
However, they perform significantly better with the SimCLR initializer on image datasets and can even outperform previous state-of-the-art methods for learning under label noise.
Even the typical method, i.e., training a deep neural network-based classifier under the categorical cross-entropy loss, can outperform previous state-of-the-art methods under some noise conditions. 
These observations suggest that lack of good visual representations is a possible reason that many supervised robust methods perform poorly on image classification tasks. 
We believe that our findings can serve as a new baseline for learning under label noise on image datasets. 
Moreover, we believe that decoupling the representation learning problem from learning under label noise would lead to new methods that can do well on either of these tasks with complementary strengths without the need for methods targeting both of these tasks together.

{\small
\bibliographystyle{ieee_fullname}
\bibliography{egbib}

\begin{thebibliography}{10}\itemsep=-1pt

\bibitem{m-correction}
Eric Arazo, Diego Ortego, Paul Albert, Noel O’Connor, and Kevin McGuinness.
\newblock Unsupervised label noise modeling and loss correction.
\newblock In {\em ICML}, pages 312--321. PMLR, 2019.

\bibitem{mixmatch}
David Berthelot, Nicholas Carlini, Ian Goodfellow, Nicolas Papernot, Avital
  Oliver, and Colin Raffel.
\newblock Mixmatch: A holistic approach to semi-supervised learning.
\newblock {\em arXiv preprint arXiv:1905.02249}, 2019.

\bibitem{identifying}
Carla~E Brodley and Mark~A Friedl.
\newblock Identifying mislabeled training data.
\newblock {\em Journal of artificial intelligence research}, 11:131--167, 1999.

\bibitem{identifying-mislabeled}
Carla~E Brodley, Mark~A Friedl, et~al.
\newblock Identifying and eliminating mislabeled training instances.
\newblock In {\em AAAI}, pages 799--805, 1996.

\bibitem{simclr}
Ting Chen, Simon Kornblith, Mohammad Norouzi, and Geoffrey Hinton.
\newblock A simple framework for contrastive learning of visual
  representations.
\newblock In {\em ICML}, pages 1597--1607. PMLR, 2020.

\bibitem{simclrv2}
Ting Chen, Simon Kornblith, Kevin Swersky, Mohammad Norouzi, and Geoffrey
  Hinton.
\newblock Big self-supervised models are strong semi-supervised learners.
\newblock {\em arXiv preprint arXiv:2006.10029}, 2020.

\bibitem{webly}
Xinlei Chen and Abhinav Gupta.
\newblock Webly supervised learning of convolutional networks.
\newblock In {\em ICCV}, pages 1431--1439, 2015.

\bibitem{imagenet32}
Patryk Chrabaszcz, Ilya Loshchilov, and Frank Hutter.
\newblock A downsampled variant of imagenet as an alternative to the cifar
  datasets.
\newblock {\em arXiv preprint arXiv:1707.08819}, 2017.

\bibitem{c2d}
Zheltonozhskii Evgenii, Baskin Chaim, Mendelson Avi, M.~Bronstein Alex, and Or
  Litany.
\newblock Contrast to divide: self-supervised pre-training for learning with
  noisy labels.
\newblock 2021.
\newblock under review.

\bibitem{bilevel}
Luca Franceschi, Paolo Frasconi, Saverio Salzo, Riccardo Grazzi, and
  Massimiliano Pontil.
\newblock Bilevel programming for hyperparameter optimization and
  meta-learning.
\newblock In {\em ICML}, pages 1568--1577, 2018.

\bibitem{ghosh2017}
Aritra Ghosh, Himanshu Kumar, and PS Sastry.
\newblock Robust loss functions under label noise for deep neural networks.
\newblock In {\em AAAI}, pages 1919--1925, 2017.

\bibitem{robust-mwnet}
Aritra Ghosh and Andrew Lan.
\newblock Do we really need gold samples for sample weighting under label
  noise?
\newblock In {\em WACV}, pages 3922--3931, January 2021.

\bibitem{ghosh2015}
Aritra Ghosh, Naresh Manwani, and PS Sastry.
\newblock Making risk minimization tolerant to label noise.
\newblock {\em Neurocomputing}, 160:93--107, 2015.

\bibitem{ghosh2017-dt}
Aritra Ghosh, Naresh Manwani, and PS Sastry.
\newblock On the robustness of decision tree learning under label noise.
\newblock In {\em PAKDD}, pages 685--697. Springer, Cham, 2017.

\bibitem{s-model}
Jacob Goldberger and Ehud Ben-Reuven.
\newblock Training deep neural-networks using a noise adaptation layer.
\newblock In {\em ICLR}, 2017.

\bibitem{masking}
Bo Han, Jiangchao Yao, Gang Niu, Mingyuan Zhou, Ivor Tsang, Ya Zhang, and
  Masashi Sugiyama.
\newblock Masking: A new perspective of noisy supervision.
\newblock In {\em NeurIPS}, pages 5836--5846, 2018.

\bibitem{co-teaching}
Bo Han, Quanming Yao, Xingrui Yu, Gang Niu, Miao Xu, Weihua Hu, Ivor Tsang, and
  Masashi Sugiyama.
\newblock Co-teaching: Robust training of deep neural networks with extremely
  noisy labels.
\newblock In {\em NeurIPS}, pages 8527--8537, 2018.

\bibitem{pretraining}
Dan Hendrycks, Kimin Lee, and Mantas Mazeika.
\newblock Using pre-training can improve model robustness and uncertainty.
\newblock In {\em ICML}, pages 2712--2721. PMLR, 2019.

\bibitem{self-supervision}
Dan Hendrycks, Mantas Mazeika, Saurav Kadavath, and Dawn Song.
\newblock Using self-supervised learning can improve model robustness and
  uncertainty.
\newblock {\em arXiv preprint arXiv:1906.12340}, 2019.

\bibitem{mentornet}
Lu Jiang, Zhengyuan Zhou, Thomas Leung, Li-Jia Li, and Li Fei-Fei.
\newblock Mentornet: Learning data-driven curriculum for very deep neural
  networks on corrupted labels.
\newblock In {\em ICML}, pages 2304--2313, 2018.

\bibitem{dividemix}
Junnan Li, Richard Socher, and Steven~CH Hoi.
\newblock Dividemix: Learning with noisy labels as semi-supervised learning.
\newblock {\em arXiv preprint arXiv:2002.07394}, 2020.

\bibitem{l2l-noise}
Junnan Li, Yongkang Wong, Qi Zhao, and Mohan~S Kankanhalli.
\newblock Learning to learn from noisy labeled data.
\newblock In {\em CVPR}, pages 5051--5059, 2019.

\bibitem{focal}
Tsung-Yi Lin, Priya Goyal, Ross Girshick, Kaiming He, and Piotr Doll{\'a}r.
\newblock Focal loss for dense object detection.
\newblock In {\em ICCV}, pages 2980--2988, 2017.

\bibitem{elr}
Sheng Liu, Jonathan Niles-Weed, Narges Razavian, and Carlos Fernandez-Granda.
\newblock Early-learning regularization prevents memorization of noisy labels.
\newblock {\em arXiv preprint arXiv:2007.00151}, 2020.

\bibitem{peer}
Yang Liu and Hongyi Guo.
\newblock Peer loss functions: Learning from noisy labels without knowing noise
  rates.
\newblock In {\em ICML}, 2020.

\bibitem{normalized-loss}
Xingjun Ma, Hanxun Huang, Yisen Wang, Simone Romano, Sarah Erfani, and James
  Bailey.
\newblock Normalized loss functions for deep learning with noisy labels.
\newblock In {\em ICML}, 2020.

\bibitem{forward}
Giorgio Patrini, Alessandro Rozza, Aditya Krishna~Menon, Richard Nock, and
  Lizhen Qu.
\newblock Making deep neural networks robust to label noise: A loss correction
  approach.
\newblock In {\em CVPR}, pages 1944--1952, 2017.

\bibitem{reed}
Scott Reed, Honglak Lee, Dragomir Anguelov, Christian Szegedy, Dumitru Erhan,
  and Andrew Rabinovich.
\newblock Training deep neural networks on noisy labels with bootstrapping.
\newblock {\em ICLR Workshops}, 2015.

\bibitem{l2rw}
Mengye Ren, Wenyuan Zeng, Bin Yang, and Raquel Urtasun.
\newblock Learning to reweight examples for robust deep learning.
\newblock In {\em ICML}, pages 4334--4343, 2018.

\bibitem{mwnet}
Jun Shu, Qi Xie, Lixuan Yi, Qian Zhao, Sanping Zhou, Zongben Xu, and Deyu Meng.
\newblock Meta-weight-net: Learning an explicit mapping for sample weighting.
\newblock In {\em NeurIPS}, pages 1919--1930, 2019.

\bibitem{joint}
Daiki Tanaka, Daiki Ikami, Toshihiko Yamasaki, and Kiyoharu Aizawa.
\newblock Joint optimization framework for learning with noisy labels.
\newblock In {\em CVPR}, pages 5552--5560, 2018.

\bibitem{unhinged}
Brendan Van~Rooyen, Aditya Menon, and Robert~C Williamson.
\newblock Learning with symmetric label noise: The importance of being
  unhinged.
\newblock In {\em NeurIPS}, pages 10--18, 2015.

\bibitem{veit2017learning}
Andreas Veit, Neil Alldrin, Gal Chechik, Ivan Krasin, Abhinav Gupta, and Serge
  Belongie.
\newblock Learning from noisy large-scale datasets with minimal supervision.
\newblock In {\em CVPR}, pages 839--847, 2017.

\bibitem{xiao2015learning}
Tong Xiao, Tian Xia, Yi Yang, Chang Huang, and Xiaogang Wang.
\newblock Learning from massive noisy labeled data for image classification.
\newblock In {\em CVPR}, pages 2691--2699, 2015.

\bibitem{ldmi}
Yilun Xu, Peng Cao, Yuqing Kong, and Yizhou Wang.
\newblock L\_dmi: An information-theoretic noise-robust loss function.
\newblock 32:6225--6236, 2019.

\bibitem{lars}
Yang You, Igor Gitman, and Boris Ginsburg.
\newblock Large batch training of convolutional networks.
\newblock {\em arXiv preprint arXiv:1708.03888}, 2017.

\bibitem{yuan2018iterative}
Bodi Yuan, Jianyu Chen, Weidong Zhang, Hung-Shuo Tai, and Sara McMains.
\newblock Iterative cross learning on noisy labels.
\newblock In {\em WACV}, pages 757--765. IEEE, 2018.

\bibitem{mixup}
Hongyi Zhang, Moustapha Cisse, Yann~N Dauphin, and David Lopez-Paz.
\newblock mixup: Beyond empirical risk minimization.
\newblock {\em arXiv preprint arXiv:1710.09412}, 2017.

\bibitem{generalized-ce}
Zhilu Zhang and Mert Sabuncu.
\newblock Generalized cross entropy loss for training deep neural networks with
  noisy labels.
\newblock In {\em NeurIPS}, pages 8778--8788, 2018.

\end{thebibliography}
}

\end{document}